\documentclass{cys}
 \usepackage[square,numbers]{natbib} 
\usepackage[utf8]{inputenc}
\usepackage[english]{babel}
\usepackage{graphicx}
\usepackage{epstopdf} 
\usepackage[nolist,nohyperlinks]{acronym}
\usepackage{tabularx}
\usepackage{booktabs}
\usepackage{arydshln}
\usepackage{changepage}
\usepackage{multirow}
\usepackage{amssymb,amsmath}
\usepackage{url}

\addto\captionsenglish{%

}

\addto\captionsspanish{%

}

\begin{acronym}
  \acro{MSC}{Multi-Sentence Compression}
  \acro{WG}{Word Graph}
  \acro{POS}{Part-of-Speech}
  \acro{NLP}{Natural Language Processing}
  \acro{ATS}{Automatic Text Summarization}
  \acro{LDA}{Latent Dirichlet Allocation}
  \acro{LSI}{Latent Semantic Indexing}
  \acro{CR}{Compression Ratio}
  \acro{ILP}{Integer Linear Programming}
  \acro{TSP}{Traveling Salesman Problem}
  \acro{NN}{Neural Network}
  \acro{SC}{Sentence Compression}
  \acro{WG}{Word Graph}
\end{acronym}

\title{A Multilingual Study of Multi-Sentence Compression using Word Vertex-Labeled Graphs and Integer Linear Programming}

\author{Elvys Linhares Pontes$^{1,2,3}$, Stéphane Huet$^2$, Juan-Manuel Torres-Moreno$^{2,3}$, \\ 
Thiago G. da Silva$^{4,5}$, Andréa Carneiro Linhares$^6$}

\affil{ 
$^1$ L3i, University of La Rochelle, La Rochelle, France \authorcr   
$^2$ LIA, University of Avignon, Avignon, France \authorcr
$^3$ Polytechnique Montréal, Montréal, Canada \authorcr
$^4$ Inst. Federal de Educação, Ciência e Tecnologia da Paraiba, PB, Brazil \authorcr
$^5$ Instituto de Computação Univ. Federal Fluminense, RJ, Brazil \authorcr
$^6$ Universidade Federal do Ceará, Sobral, Brazil 
\authorcr  \authorcr
elvys.linhares\_pontes@univ-lr.fr
\authorcr  \authorcr
}

\begin{document}

\maketitle

\renewcommand{\tablename}{Table}

\begin{abstract}
\ac{MSC} aims to generate a short sentence with the key information from a cluster of similar sentences.
MSC enables summarization and question-answering systems to generate outputs combining fully formed sentences from one or several documents.
This paper describes an Integer Linear Programming method for \ac{MSC} using a vertex-labeled graph to select different keywords, with the goal of generating more informative sentences while maintaining their grammaticality. 
Our system is of good quality and outperforms the state of the art for evaluations led on news datasets in three languages: French, Portuguese and Spanish. 
We led both automatic and manual evaluations to determine the informativeness and the grammaticality of compressions for each dataset.
In additional tests, which take advantage of the fact that the length of compressions can be modulated, we still improve ROUGE scores with shorter output sentences.
\end{abstract}

\section{Introduction}
\label{sc:int}

A considerable amount of information is published in various sites every day, e.g. comments, photos, videos and audio in different languages.
The increased number of electronic devices (smartphones, tablets, etc.) have made access to these information easier and faster.
Moreover, websites such as Wikipedia or news aggregators can provide detailed data on various issues but texts may be long and convey a lot of information.
Readers, besides not having the time to go through this amount of information, are not interested in all the proposed subjects and generally select the content of their interest.
One solution to this problem is the generation of summaries containing only the key information.

Among the various applications of \ac{NLP}, \ac{ATS} aims to automatically identify the relevant data inside one or more documents, and create a condensed text with the main information~\cite{LINHARESPONTES2020101763}. At the same time, summaries should be short with as little redundant information as possible. Summarization systems usually rely on statistical, morphological and syntactic analysis approaches~\cite{torres:2014}.
Some of them use \acf{MSC} in order to produce from a set of similar sentences a small-sized sentence which is both grammatically correct and informative~\cite{banerjee:2015,filippova:2010,LINHARESPONTES2020101763}. 
Although compression is a challenging task, it is appropriate to generate summaries that are more informative than the state-of-the-art extractive methods for \ac{ATS}.

The contributions of this article are two-fold. (i) We improved 
the model for \ac{MSC}~\cite{elvys:2018:textgraph} that extends the common approach based on Graph Theory, using vertex-labeled graphs and \ac{ILP} to select the best compression. The vertex-labeled graphs\footnote{A vertex-labeled graph means a graph where each node has a label. In this work, a label is represented by a color and different nodes can have the same label.} are used to model a cluster of similar sentences with keywords. (ii) Whereas previous work usually limited the experimental study on one or two datasets, we tested our model on three corpora, each in a different language. Evaluations led with both automatic metrics and human evaluations show that our ILP model consistently generate more informative sentences than two state-of-the-art systems while maintaining their grammaticality. Interestingly, our approach is able to choose the amount of information to keep in the compression output, through the definition of the maximum compression length.

This paper is organized as follows: we describe and survey the \ac{MSC} problem in Section~\ref{sc:msc}.
Next, we detail our approach in Section~\ref{sc:oa}.
The experiments and the results are discussed in Sections~\ref{sc:exp} and~\ref{sc:res}.
Lastly, conclusions and some final comments are set out in Section~\ref{sc:conc}.
\section{Related Work}
\label{sc:msc}

\ac{SC} aims at producing a reduced grammatically correct sentence. Compressions may have different \ac{CR} levels,\footnote{The \ac{CR} is the length of the compression divided by the average length of all source sentences} whereby the lower the \ac{CR} level, the higher the reduction of the information is.
SC can be employed in the contexts of the summarization of documents, the generation of article titles or the simplification of complex sentences, using diverse methods such as optimization~\cite{clarke:2007,Clarke:2008}, syntactic analysis, deletion of words~\cite{filippova:2015} or generation of sentences~\cite{MiaoB16,RushCW15}.
Recently, many \ac{SC} approaches using \ac{NN} have been developed~\cite{MiaoB16,RushCW15}. 
These methods may generate good results for a single sentence because they combine many complex structures such as recurrent neural networks (based on Gated Recurrent Units and Long Short Term Memory), the sequence-to-sequence paradigm and condition mechanisms (e.g., attention). 
However, these composite neural networks need huge corpora to learn how to generate compressions (e.g., Rush et al. used the Gigaword corpus that contains around 9.5 million news) and take a lot of time to accomplish the learning process.

\acf{MSC}, also coined as Multi-Sentence Fusion, is a variation of SC. 
Unlike SC, \ac{MSC} combines the information of a cluster of similar sentences to generate a new sentence, hopefully grammatically correct, which compresses the most relevant data of this cluster.
The idea of \ac{MSC} was introduced by Barzilay and McKeown~\cite{barzilay:2005}, who developed a multi-document summarizer which represents each sentence as a dependency tree; their approach aligns and combines these trees to fusion sentences.
Filippova and Strube~\cite{filippova:2008} also used dependency trees to align each cluster of related sentences and generated a new tree, this time with \ac{ILP}, to compress the information.
In 2010, Filippova presented a new model for \ac{MSC}, simple but effective, which is based on Graph Theory and a list of stopwords.
She used a \ac{WG} to represent and to compress a cluster of related sentences; the details of this model, which is extended by the work of this paper, can be found in Section~\ref{ssc:fil}.

Inspired by the good results of the Filippova's method, many studies have used it in a first step to generate a list of the $N$ shortest paths, then have relied on different reranking strategies to analyze the candidates and select the best compression~\cite{banerjee:2015,boudin:2013,luong:2015,tzouridis:2014}.
Boudin and Morin~\cite{boudin:2013} developed a reranking method measuring the relevance of a candidate compression using \textit{key phrases}\footnote{\textit{key phrases} are words that capture the main topics of a document.}, obtained with the TextRank algorithm~\cite{mihalcea:2004}, and the length of the sentence. 
Another reranking strategy was proposed by Luong et al.~\cite{luong:2015}.
Their method ranks the sentences from the counts of unigrams\footnote{An $n$-gram is a contiguous sequence of $n$ items from a given text.} occurring in every source sentence.
ShafieiBavani et al.~\cite{ShafieiBavani:2016} also used a \ac{WG} model; their approach consists of three main components:
(i) a merging stage based on Multiword Expressions (MWE), (ii) a mapping strategy
based on synonymy between words and (iii) a reranking step to identify the best compression
candidates generated using a Part-of-Speech-based language model (POS-LM).
Tzouridis et al.~\cite{tzouridis:2014} proposed a structured learning-based approach. Instead of applying heuristics as Filippova~\cite{filippova:2010}, they adapted the decoding process to the data by parameterizing a shortest path algorithm. They devised a structural support vector machine to learn the shortest path in possibly high dimensional joint feature spaces and proposed a generalized loss-augmented decoding algorithm that is solved exactly by \ac{ILP} in polynomial time.

Linhares Pontes et al.~\cite{elvys:2018:textgraph} also presented an \ac{ILP} approach that models a set of similar sentences as vertex-labeled word graphs.  Their approach selects keywords and relevant 3-grams to generate more informative compressions while maintaining their grammaticality as possible. They have studied the quality of compressions by analyzing different amounts of keywords in order to manage both the length and the informativeness of compressions.

We found two other studies that applied \ac{ILP} to combine and compress several sentences. Banerjee et al.~\cite{banerjee:2015} developed a multi-document \ac{ATS} system that generated summaries after compressing similar sentences.
They used Filippova's method to generate 200 random compressed sentences.
Then they created an \ac{ILP} model to select the most informative and grammatically correct compression.
Thadani and McKeown~\cite{Thadani:2013} proposed another \ac{ILP} model using an inference approach for sentence fusion. Their \ac{ILP} formulation relies on n-gram factorization and aims at avoiding cycles and disconnected structures.

In the ATS task, Shang et al.~\cite{shang-etal-2018-unsupervised} adapted the Boudin and Morin's approach~\cite{boudin:2013} to take into account the grammaticality for the reranking of compressions. Instead of the TextRank algorithm, they analyze the spreading influence in WG to generate more informative and grammatical compressions and to improve the quality of summaries. Nayeem et al.~\cite{nayeem-etal-2018-abstractive} designed a paraphrastic sentence fusion model which jointly performs sentence fusion and paraphrasing using skip-gram word embedding model at the sentence level.

Recently, Zhao et al.~\cite{zhao-etal-2019-unsupervised} presented an unsupervised rewriter to improve the grammaticality of MSC outputs while introducing new words. They used the WG approach to produce coarse-grained compressions, from which they substitute words with their shorter synonyms to yield paraphrased sentence. Then, their neural rewriter proposes paraphrases for these compressions in order to improve grammaticality and encourage more novel words. 

Another related task is the sentence aggregation that combines a group of sentences, not necessarily with a similar semantic content, to generate a single sentence (e.g., ``\textit{The car is here.}'' and ``\textit{It is blue.}'' can be aggregated into ``\textit{The blue car is here.}''). This aggregation can be at semantic and syntactic levels~\cite{ReapeMellish99}. 
The aggregation rules can be acquired automatically from  a corpus~\cite{Barzilay:2006}.
However, this process is not possible for all situations and the sentence aggregation depends on the sentence planning to combine the sentences.

Following previous studies for MSC that rely on Graph Theory with good results, this work presents a new \ac{ILP} framework that takes into account keywords for \ac{MSC}.
We compare our learning approach to the graph-based sentence compression techniques proposed by Filippova~\cite{filippova:2010} and Boudin and Morin~\cite{boudin:2013}, considered as state-of-the-art methods for \ac{MSC}. We intend to apply our method on various languages and not to be dependent on linguistic resources or tools specific to languages.  This led us to put aside systems which, despite being competitive, rely on resources like WordNet or Multiword expression detectors~\cite{ShafieiBavani:2016}.
Since we borrowed concepts and ideas from Filippova's method, we detail her approach in the next section.

\subsection{Filippova's Method}
\label{ssc:fil}

\begin{figure*}[h]
  \centering
  \includegraphics[width=12cm]{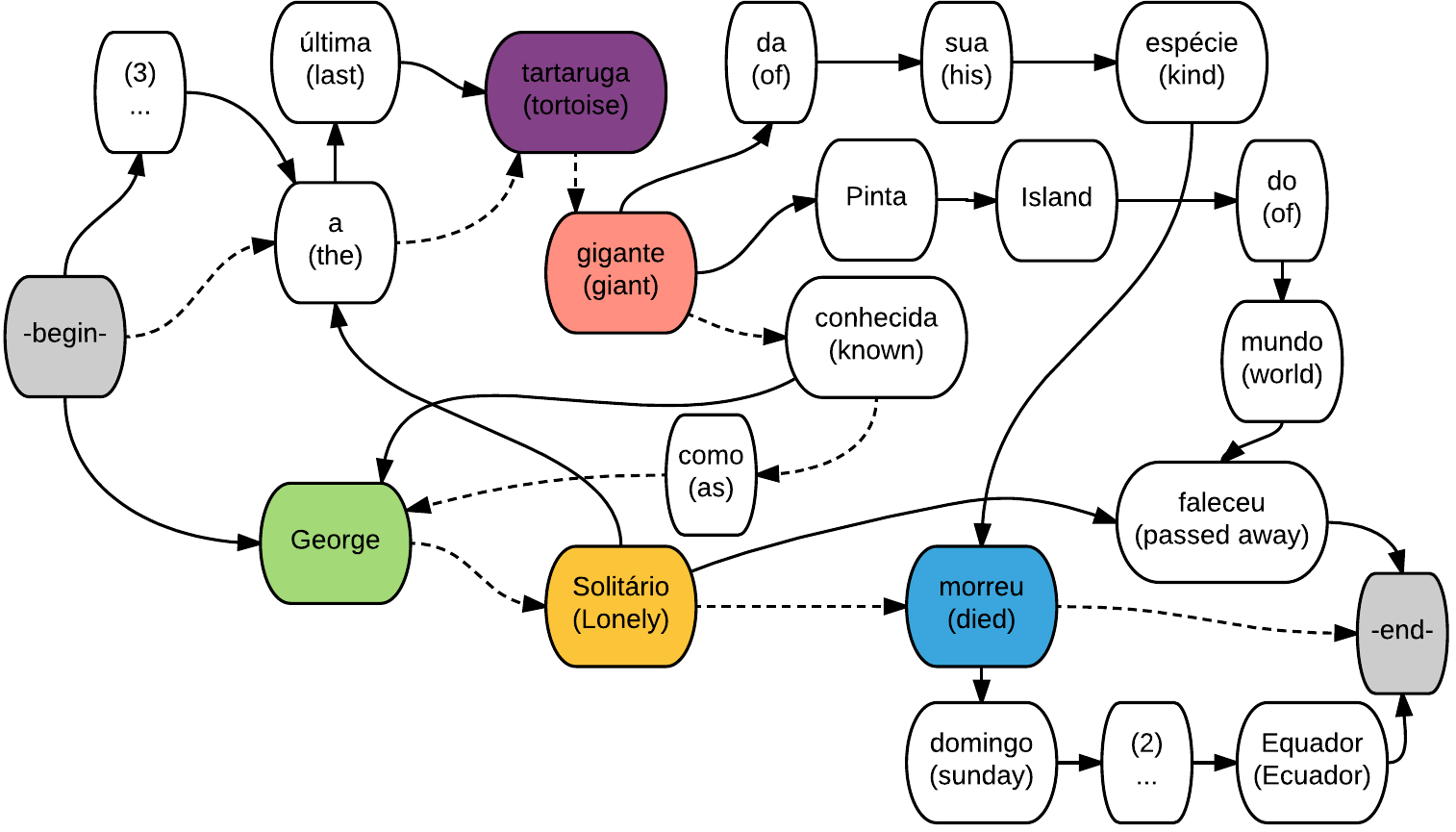}
  \caption {WG generated from the sentences (1) to (4) (without the punctuation and \ac{POS} for easy readability). The dotted path represents the best compression for this \ac{WG} and the colored vertices represent the keywords of the document.}
  \label{img:wg}
\end{figure*}

Filippova~\cite{filippova:2010} modeled a document $D$ containing $n$ similar sentences $\{s_1, s_2, \ldots, s_n \}$, as a directed word graph $G=(V,A)$. 
$V$ is the set of vertices (words) and $A$ is the set of arcs (adjacency relationship).
Figure \ref{img:wg} illustrates the word graph of the following Portuguese sentences:

\begin{enumerate}
	\item {\it George Solitário, a última tartaruga gigante Pinta Island do mundo, faleceu.} {\small (Lonesome George, the world’s last Pinta Island
giant tortoise, has passed away.)}
    \item {\it A tartaruga gigante conhecida como George Solitário morreu domingo no Parque Nacional de Galapagos, Equador.} {\small(The giant tortoise known as Lonesome George died Sunday at the Galapagos National Park in Ecuador.)}
    \item {\it Ele tinha apenas cem anos de vida, mas a última tartaruga gigante Pinta  conhecida, George Solitário, faleceu.} {\small(He was only about a hundred years old, but the last known giant Pinta tortoise, Lonesome George, has passed away.)}
    \item {\it George Solitário, a última tartaruga gigante da sua espécie, morreu.} {\small(Lonesome George, a giant tortoise believed to be the last of his kind, has died.)}
\end{enumerate}

The initial graph $G$ is composed of the first sentence (1) and the vertices --begin-- and --end--.
For a new sentence, a new vertex is created when a word/\ac{POS} pair cannot be matched to an existing vertex of $G$ once lowercased. Besides, at most one occurrence of a given word/\ac{POS} inside a sentence can be associated with a given vertex.

Sentences are individually analyzed and added to $G$. 
Each sentence represents a simple path between the --begin-- and --end-- vertices and its words are inserted in the following order:

\begin {enumerate}
\item Non-stopwords for which no candidate exists in the graph or for which an unambiguous mapping is possible;
\item Non-stopwords for which there are several possible candidates in the graph that may occur more than once in the sentence;
\item Stopwords.
\end {enumerate}

In cases 2 and 3, the word mapping is ambiguous because there is more than one vertex in the graph that references the same word/POS.
In this case, we analyze the immediate context (the preceding and following words/POSs in the sentence and the neighboring nodes in the graph) or the frequency (i.e., the number of words that were mapped to the considered vertex) to select the best candidate node.

Once vertices have been added, arcs are valued by weights which represent the levels of cohesion between two words in the graph (Equation~\ref{eq:1}). 
Cohesion is calculated from the frequency and the position of these words in sentences, according to  Equation~\ref{eq:2}:

\begin{equation}
w(i,j) = \frac{\textrm{cohesion}(i,j)}{\textrm{freq}(i) \times \textrm{freq}(j)},
\label{eq:1}
\end{equation}

\begin{equation}
\textrm{cohesion}(i,j) = \frac{\textrm{freq}(i) + \textrm{freq}(j)}{\sum_{s \in D} \textrm{diff}(s,i,j)^{-1}},
\label{eq:2}
\end{equation}

\noindent where $\textrm{freq}(i)$ is the word frequency mapped to the vertex $i$ and the function $\textrm{diff}(s,i,j)$ refers to the distance between the offset positions of words $i$ and $j$ in the sentences $s$ of $D$ containing these two words.

From the graph $G$, the system calculates the 50 shortest paths that are longer than eight words and have at least one verb.
Finally, the system reranks the paths by normalizing the total path weight over their length and selects the path with the lowest score as the best \ac{MSC}.
\section{Our Approach}
\label{sc:oa}

Filippova's method chooses the path with the lowest score taking into account the level of cohesion between two adjacent words in the document.
However, two words with a strong cohesion do not necessarily have a good informativeness because the cohesion only measures the distance and the frequency of words in the sentences.
In this work, we propose a method to concurrently analyze cohesion and keywords in order to generate a more informative and comprehensible compression.

Our method calculates the shortest path from the cohesion of words and grants bonuses to the paths that have different keywords.
For this purpose, our approach is based on Filippova's method (Section~\ref{ssc:fil}) to model a document $D$ as a graph and to calculate the cohesion of words.
In addition, we analyze the keywords of the document to favor hypotheses with meaningful information.

\subsection{Keyword Extraction}

Introducing keywords in the graph helps the system to generate more informative compressions because it takes into account the words that are representative of the cluster to calculate the best path in the graph, and not only the cohesion and frequency of words.
Keywords can be identified for each cluster with various extraction methods and we study three widely used techniques: \ac{LSI}, \ac{LDA} and TextRank. Despite the small number of sentences per cluster, these methods generate good results because clusters are composed of similar sentences with a high level of redundancy.
\ac{LSI} uses Singular-Value Decomposition (SVD), a technique closely related to eigenvector decomposition and factor analysis, to model the associative relationships~\cite{Deerwester:1990}.
\ac{LDA} is a topic model that generates topics based on word frequency from a set of documents~\cite{Blei:2003}.
Finally, TextRank algorithm analyzes the words in texts using \ac{WG}s and estimates their relevance~\cite{mihalcea:2004}. For \ac{LDA} whose modeling is based on the concept of topics, we consider that the document $D$ describes only one topic since it is composed of semantically close sentences related to a specific news item.
A same word or keyword can be represented by one or several nodes in \ac{WG}s (see Section \ref{ssc:fil}). 
In order to prioritize the sentence generation containing multiple keywords and to reduce the redundancy, we add a bonus to the compression score when the compression contains different keywords.

\subsection{Vertex-Labeled Graph}
\label{sc:vlg}

A vertex-labeled graph is a graph $G = (V, A)$ with a label on the vertices $K = \{0, ... , |K|\}$, where $|K|$ is the number of different labels.
This graph type has been employed in several domains such as biology~\cite{zheng:2011} or \ac{NLP}~\cite{Bruckner:2013}. In this last study, the correction of Wikipedia inter-language links was modeled as a Colorful Components problem. Given a vertex-colored graph, the Colorful Components problem aims at finding the minimum-size edge sets that are connected and do not have two vertices with the same color.

In the context of \ac{MSC}, we want to generate a short informative compression where keyword may be represented by several nodes in the word graph. Labels enable us to represent keywords in vertex-labeled graphs and generate a compression without repeated keywords while preserving the informativeness.
In this framework, we grant bonuses only once for nodes with the same label to prioritize new information in the compression (Figure \ref{img:wg}).
To make our model coherent, we added a base label (label 0) for all non-keywords in the word graph.
The following section describes our \ac{ILP} model to select sentences including labeled keywords inside \ac{WG}s.

\subsection{\ac{ILP} Modeling}
\label{sc:mm}

There are several algorithms with a polynomial complexity to find the shortest path in a graph.
However, the restriction on the minimum number $P_\textrm{min}$ of vertices (i.e., the minimum number of words in the compression) makes the problem NP-hard. Indeed, let $v_0$ be the --begin-- vertex. If $P_\textrm{min}$ equals $|V|$ and if we add an auxiliary arc from --end-- vertex to $v_0$, our problem is similar to the \ac{TSP}, which is NP-hard.

For this work we use the formulation known as Miller-Tucker-Zemlin (MTZ) to solve our problem~\cite{oncan:2009,Thadani:2013}.
This formulation uses a set of auxiliary variables, one for each vertex in order to prevent a vertex from being visited more than once in the cycle and a set of arc restrictions.

The problem of production of a compression that favors  informativeness and grammaticality is expressed  as Equation~\ref{fo}.
In other words, we look for a path (sentence) that has a good cohesion and contains a maximum of labels (keywords).

\begin{equation}
  \mathrm{Minimize} ~ 
	\Big( \sum_{(i,j) \in A} w(i,j) \cdot x_{i,j}  - c \cdot \sum_{k \in K} b_k
    \Big) \label{fo}
\end{equation}

\noindent where $x_{ij}$ indicates the existence of the arc $ (i, j) $ in the solution, $w(i,j)$ is the cohesion of the words $i$ and $j$ (Equation~\ref{eq:1}), 
$K$ is the set of labels (each representing a keyword), $b_k$ indicates the existence of a word with label (keyword) $k$ in the solution and $c$ is the keyword bonus of the graph.\footnote{The keyword bonus allows the generation of longer compressions that may be more informative.}

\subsection{Structural Constraints}
\label{sc:sc}

We describe the structural constraints for the problem of consistency in compressions and define the bounds of the variables. 
First, we consider the problem of consistency which requires an inner and an outer arc active for every word used in the solution, where $y_v$ indicates the existence of the vertex $v$ in the solution.

\begin{align}
\sum_{i\in\delta^+(v)} x_{vi} = y_v &&\forall v\in V, \label{eq5} \\
\sum_{i\in\delta^-(v)} x_{iv} = y_v &&\forall v\in V. \label{eq4} 
\end{align}

The constraints (\ref{eq1}) and (\ref{eq1.1}) control the minimum and the maximum number of vertices ($P_\textrm{min}$ and $P_\textrm{max}$) used in the solution respectively, i.e., the minimum and the maximum number of words in the final compression.

\begin{align}
\sum_{v \in V} y_v \geq P_\textrm{min}, \label{eq1}
\end{align}

\begin{align}
\sum_{v \in V} y_v \leq P_\textrm{max}. \label{eq1.1}
\end{align}
 
The set of constraints (\ref{eq2}) matches label variables (keywords) with vertices (words), where $V(k)$ is the set of all vertices with label $k$.

\begin{align}
\sum_{v \in V(k)} y_v \geq b_k, && \forall k \in K. \label{eq2}
\end{align}

Equality (\ref{eq6}) sets the vertex $v_0$ in the solution.

\begin{align}
y_0 = 1. \label{eq6}
\end{align}

The restrictions (\ref{eq7}) and (\ref{eq8}) are responsible for the elimination of sub-cycles, where $u_v$ ($\forall v \in V$) are auxiliary variables for the elimination of sub-cycles and $M$ is a large number (e.g., $M = |V|$).

\begin{align}
u_0 = 1, \label{eq7} \\
u_i - u_j +1 \leq M - M \cdot x_{ij} && \forall (i,j)\in A, j \neq 0. \label{eq8} 
\end{align}

Finally, equations (\ref{eq9}) -- (\ref{eq13}) define the field of variables.

\begin{align}
x_{ij} \in \{ 0,1 \}, && \forall (i,j) \in A, \label{eq9} \\
y_{v} \in \{ 0,1 \}, && \forall v \in V, \label{eq11} \\
u_{v} \in \{ 1, 2, \ldots, |V| \}, && \forall v \in V. \label{eq13}
\end{align}

We calculate the 50 best solutions according to the objective (\ref{fo}) having at least eight words and at least one verb.
Specifically, we find the best solution, then we add a constraint in the model to avoid this solution and repeat this process 50 times to find the other solutions.

The optimized score (Equation~\ref{fo}) explicitly takes into account the size of the generated sentence.
Contrary to Filippova's method, sentences may have a negative score because we subtract from the cohesion value of the path the introduced scores for keywords.
Therefore, we use the exponential function to ensure a score greater than zero.
Finally, we select the sentence with the lowest final score (Equation~\ref{eq:snorm}) as the best compression.

\begin{equation} \label{eq:snorm}
   \textrm{score}_{norm}(s) = \frac{e^{\textrm{score}_{opt}(s)} }{||s||},
\end{equation}

\noindent where $\textrm{score}_{opt}(s)$ is the score of the sentence $s$ from Equation \ref{fo}. 
\section{Experimental Setup}
\label{sc:exp}

Algorithms were implemented using the Python programming language with the \textsf{takahe}\footnote {\url{http://www.florianboudin.org/publications.html}} and \textsf{gensim}\footnote{\url{https://radimrehurek.com/gensim/models/ldamodel.html}} libraries.
The mathematical model was implemented in C++ with the \textsf{Concert} library and we used the solver \textsf{CPLEX 12.6}\footnote{\url{https://www.ibm.com/products/ilog-cplex-optimization-studio}}.

The objective function (see Equation~\ref{fo}) involves a keyword bonus. Since each WG can have weight arcs of different values, fixing this bonus is decisive to allow the generation of slightly longer compressions. We tested several metrics (fixed values, the arithmetic average, the median, and the geometric average of the weights arcs of WG) to define the keyword bonus of the WG and empirically found that geometric mean outperformed others.

\subsection{Evaluation Datasets}

Various corpora have been developed for MSC and are composed of clusters of similar sentences from different source news in English, French, Portuguese, Spanish or Vietnamese languages.
Whereas the data built by McKeown et al.~\cite{McKeown:2010} and Luong et al.~\cite{luong:2015} have clusters limited to pairs of sentences,
the corpora made by Filippova~\cite{filippova:2010}, Boudin and Morin~\cite{boudin:2013}, and Linhares Pontes et al.~\cite{elvys:lrec} contain clusters of at least 7 similar sentences. 
McKeown et al.~\cite{McKeown:2010} collected 300 English sentence pairs taken from newswire clusters using Amazon’s Mechanical Turk, while the corpus introduced in Luong et al.~\cite{luong:2015} is made of 250 Vietnamese sentences divided into 115 groups of similar sentences with 2 sentences by group. 
McKeown et al.~\cite{McKeown:2010}, Luong et al.~\cite{luong:2015}, Boudin and Morin~\cite{boudin:2013}, and Linhares Pontes et al.~\cite{elvys:lrec} made their corpora publicly available, but only the data associated with these last two articles are more suited to the multi-document summarization or question-answering tasks because the documents to analyze are usually composed of many similar sentences.
Therefore, we use these two corpora made of French, Portuguese and Spanish sentences.

Table~\ref{tab:statistics} summarizes the statistics of this set of data having 40 clusters of sentences for each language.
The Type-Token Ratio (TTR) indicates the reuse of tokens in a cluster and is defined by the number of unique tokens divided by the number of tokens in each cluster;
the lower the TTR, the greater the reuse of tokens in the cluster.
The sentence similarity represents the average cosine similarity of the sentences in a cluster.\footnote{The cosine similarity between two vectors $u$ and $v$ associated with two sentences is defined by $\frac{u \cdot v}{||u||\ ||v||}$ in the [0,1] range.}

The French corpus has 3 sentences compressed by native speakers for each cluster,
references having a compression rate (CR) of 60\%.
Like the French corpus, the Portuguese and Spanish corpora are composed of the first sentences of the articles found in Google News.
Each cluster is composed of related sentences and was chosen among the first sentence from different articles about Science, Sport, Economy, Health, Business, Technology, Accidents/Catastrophes, General Information and other subjects.
A cluster has at least 10 similar sentences by topic and 2 reference compressions made by different native speakers.
The average CRs are 54\% and 61\% for the Portuguese and the Spanish corpora, respectively.

\begin{table*}[t]
  \caption{\label{tab:statistics}Statistics of the corpora.}
  \begin{center}
  \begin{tabular}{lcccccc}
	\toprule
    \multirow{2}{*}{\centering \bf Characteristics} & \multicolumn{2}{c}{\bf French} & \multicolumn{2}{c}{\bf Portuguese} & \multicolumn{2}{c}{\bf Spanish}\\ 
 & \bf \small Source & \bf \small References & \bf \small Source & \bf \small References & \bf \small Source & \bf \small References \\ \cmidrule{1-7}
    \#tokens             & 20,224 & 2,362 & 17,998 & 1,425 & 30,588 & 3,694 \\
\#vocabulary (tokens) & 2,867 & 636 & 2,438 & 533 & 4,390 & 881 \\
\#sentences          & 618 & 120 & 544 & 80 & 800 & 160 \\
avg. sentence length (tokens) & 33.0 & 19.7 & 33.1 & 17.8 & 38.2 & 23.1\\
type-token ratio (TTR)    & 39\% & 50\% & 34\% & 68\% & 35\% & 43\% \\
sentence similarity & 0.46 & 0.67 & 0.51 & 0.59 & 0.47 & 0.64 \\
compression rate    &  --- &   60\%   &  --- &   54\%   & ---  &   61\%   \\
	\bottomrule
  \end{tabular}
  \end{center}
\end{table*}

The three languages derive from Latin and are closely related languages. However, they differ in many details of their grammar and lexicon.
Moreover, the datasets produced for the three languages are unlike according to several features.
First, the corpus made by Linhares Pontes et al.~\cite{elvys:lrec} contains a smaller (Portuguese corpus) and a larger (Spanish corpus) dataset in terms of sentences than the French corpus.
Besides, the compression rates of the three datasets indicate that the Portuguese source sentences have more irrelevant tokens.
The sentence similarity (Table~\ref{tab:statistics}, second last line) describes the variability of sentences in the source sentences and in the references, and reflects here that the sentences are slightly more diverse for the French corpus. This translates into a higher TTR observed for the French part (38.8\%) than for the two other languages (33.7\% and 35.2\%).

\subsection{Automatic and Manual Evaluations}

The most important features of \ac{MSC} are informativeness and grammaticality.
Informativeness measures how informational is the generated text.
As references are assumed to contain the key information, we calculated informativeness scores counting the n-grams in common between the system output and the reference compressions using ROUGE~\cite{lin:2004}.
In particular, we used the metrics ROUGE-1 and ROUGE-2, F-measure being preferred to recall for a fair comparison of various lengths of compressed sentences.
Like in~\cite{boudin:2013}, ROUGE metrics are calculated with stopwords removal and stemming\footnote{\url{http://snowball.tartarus.org/}}.

Due to limitations of the ROUGE systems that only analyze unigrams and bigrams, we also led a manual evaluation with four native speakers for French, Portuguese and Spanish.
The native speakers of each language evaluated the compression in two aspects: informativeness and grammaticality.
In the same way as Filippova~\cite{filippova:2010} as well as Boudin and Morin~\cite{boudin:2013}, the native speakers evaluated the grammaticality in a 3-point scale: 
2 points for a correct sentence; 1 point if the sentence has minor mistakes; 0 point if it is none of the above.
Like grammaticality, informativeness is evaluated in the same range: 2 points if the compression contains the main information; 1 point if the compression misses some relevant information; 0 point if the compression is not related to the main topic.
\section{Experimental Assessment}
\label{sc:res}

Compression rates are strongly correlated with human judgments of meaning and grammaticality~\cite{Napoles:2011}.
On the one hand, too short compressions may compromise sentence structure, reducing the informativeness and grammaticality.
On the other hand, longer compressions may be more interesting for \ac{ATS} when informativeness and grammaticality are decisive features.
Consequently, we analyze compression with multiple maximum compression lengths (50\%, 60\%, 70\%, 80\%, 90\% and $\infty$, the last value meaning that no constraint is fixed on the output size).

Following the idea proposed by ShafieiBavani et al.~\cite{ShafieiBavani:2016} and already implemented with success in other domains such as speech recognition~(e.g., \cite{Huet10}), we tested the use of a POS-based Language Model (POS-LM) as a post-processing stage in order to improve the grammaticality of compressions. Specifically, for each cluster, the ten best compressions according to our optimized score are reranked by a 7-gram POS-LM trained with the SRILM toolkit\footnote{\url{http://www.speech.sri.com/projects/srilm/}} on the French, Portuguese and Spanish parts of the Europarl dataset,\footnote{\url{http://www.statmt.org/europarl/}} tagged with \textsf{TreeTagger}~\cite{Schm95}.

\subsection{Results}
\label{ssec:results}

Since our method strongly depends on the set of keywords to generate informative compressions, we investigate the performance of the three keyword methods (LDA, LSI and TextRank), selecting the 5 or 10 most relevant words.
We verified the percentage of keywords generated by these methods that are included in the reference compression (Table~\ref{tb:prc_keywords}). 
A significantly higher rate of keywords in the references is observed when using LDA or LSI instead of TextRank. 
In order to obtain the most relevant words in a cluster with different sizes, we used LDA in our final \ac{MSC} system to identify 10 keywords for each cluster.

\begin{table}[htbp]
\caption{\label{tb:prc_keywords} Percentage of keywords included in the reference compression for French, Portuguese and Spanish corpora.}
\begin{center}
\begin{tabular}{lccc}
\toprule
\textbf{Methods} & \textbf{fr} & \textbf{pt} & \textbf{es}\\
\cmidrule{1-4}
LDA: 5  kws & \textbf{91\%} & \textbf{88\%}  & \textbf{85\%} \\ 
LSI: 5  kws & 90\% & 87\%  & 81\% \\ 
TextRank: 5  kws & 69\% & 55\%  & 58\%  \\ 
\cmidrule{1-4}
LDA: 10 kws & \textbf{84\%} & \textbf{70\%}  & \textbf{76\%} \\ 
LSI: 10 kws & \textbf{84\%} & 69\%  & 73\% \\ 
TextRank: 10 kws & 56\% & 44\%  & 50\%  \\ 
\bottomrule
\end{tabular}
\end{center}
\end{table}

Tables~\ref{tb:fr}, \ref{tb:pt} and~\ref{tb:es} describe the ROUGE recall scores measured for Filippova's~\cite{filippova:2010} method (named F10), Boudin and Morin's~\cite{boudin:2013} method (named BM13) and our method with multiple maximum compression lengths. As for each CR setup the size of the outputs to evaluate are comparable, the recall scores are preferred 
in this case to measure the information retained in compressions.
First, let us note that \ac{CR}s effectively observed may differ from the fixed value of $P_{max}$. For example, a 50\% threshold leads to real \ac{CR}s of 38\% to 40\% for all languages, while an 80\% level creates new sentences with real CRs between 53\% and 60\%.
Interestingly, our system obtained better ROUGE recall scores than both baselines in all languages for comparable compression lengths.
If we prioritize meaning, our method with no explicit constraint on the maximum compression length (ILP:$\infty$) improved the compression quality with a small increase of the compression length (compression ratio between 55.4\% and 65.9\%).
Instead, we can limit the length and generate compressions that are shorter and have still better ROUGE scores than the baselines.

\begin{table}[ht]
\caption{\label{tb:fr} ROUGE recall scores for multiple maximum compression lengths using the French corpus.}%
\begin{center}
\begin{tabular}{lccc}
\toprule
\multicolumn{4}{c}{\textbf{French}} \\
\textbf{Methods}      & \textbf{ROUGE-1}            & \textbf{ROUGE-2}            & \textbf{CR}    \\
\cmidrule{1-4}
F10					    & 0.5971   & 0.4072   & 51.3\%   \\
BM13			        & 0.6740   & 0.4695   & 59.8\%   \\
ILP:50\%     			& 0.4763   & 0.3039   & 39.1\%   \\
ILP:60\%        		& 0.5990   & 0.4101   & 47.4\%   \\
ILP:70\%       		    & 0.6420   & 0.4206   & 53.5\%   \\
ILP:80\%        		& 0.6783   & 0.4573   & 60.0\%   \\
ILP:90\%        		& 0.6981   & \textbf{0.4758}   & 61.8\%   \\
ILP:$\infty$    		& \textbf{0.7010}   & 0.4751   & 62.6\%   \\
\bottomrule
\end{tabular}
\end{center}
\end{table}

\begin{table}[ht]
\caption{\label{tb:pt} ROUGE recall scores for multiple maximum compression lengths using the Portuguese corpus.}%
\centering
\begin{tabular}{lccc}
\toprule
\multicolumn{4}{c}{\textbf{Portuguese}}                                \\
\textbf{Methods} & \textbf{ROUGE-1} & \textbf{ROUGE-2} & \textbf{CR}   \\
\cmidrule{1-4}
F10			     & 0.5354   & 0.2935   & 52.2\%   \\
BM13      		 & 0.6304   & 0.3493   & 69.1\%   \\
ILP:50\%         & 0.4689          & 0.2521          & 40.0\% \\
ILP:60\%         & 0.5369          & 0.2967          & 48.1\% \\
ILP:70\%         & 0.5652          & 0.3088          & 54.0\%  \\
ILP:80\%         & 0.6056          & 0.3321          & 59.0\% \\
ILP:90\%         & 0.6341          & 0.3492          & 64.6\% \\
ILP:$\infty$     & \textbf{0.6407} & \textbf{0.3546} & 65.9\% \\
\bottomrule
\end{tabular}
\end{table}

\begin{table}[ht]
\caption{\label{tb:es} ROUGE recall scores for multiple maximum compression lengths using the Spanish corpus.}%
\centering
\begin{tabular}{lccc}
\toprule
\multicolumn{4}{c}{\textbf{Spanish}}                                   \\
\textbf{Methods} & \textbf{ROUGE-1} & \textbf{ROUGE-2} & \textbf{CR}   \\
\cmidrule{1-4}
F10 		  	 & 0.4437   & 0.2631   & 43.2\%   \\
BM13      		 & 0.5167   & 0.2981   & 61.2\%   \\
ILP:50\%         & 0.3814          & 0.1990          & 38.7\% \\
ILP:60\%         & 0.4594          & 0.2651          & 45.3\% \\
ILP:70\%         & 0.5050          & 0.2922          & 50.2\% \\
ILP:80\%         & 0.5191          & 0.2982          & 53.2\% \\
ILP:90\%         & 0.5242          & 0.2982          & 54.4\% \\
ILP:$\infty$     & \textbf{0.5305} & \textbf{0.3036}          & 55.4\% \\
\bottomrule
\end{tabular}
\end{table}

Based on these results, a further analysis was done for the 80\% and $\infty$ configurations.
Table~\ref{rouge}\footnote{Although we used the same system and data as Boudin and Morin~\cite{boudin:2013} for the French corpus, we were not able to exactly reproduce their results. The ROUGE F-measure scores given in their article are close to ours for their system: 0.6568 (ROUGE-1), 0.4414 (ROUGE-2) and 0.4344 (ROUGE-SU4), but using F10 we measured higher scores than them:  0.5744 (ROUGE-1), 0.3921 (ROUGE-2) and 0.3700 (ROUGE-SU4).
} describes the results for the French, Portuguese and Spanish corpora using ROUGE F-measure scores.
The first two columns display the evaluation of the two baseline systems; the ROUGE scores measured with our method using either 80\% or $\infty$ maximum compression lengths are shown in the next two columns and the last two columns respectively. The outputs produced by all of these systems for two sample clusters in Spanish and Portuguese can be found in the Appendix.
Globally, all versions of our ILP method outperform both baselines according to ROUGE F-measures for the Portuguese and Spanish corpora, and our ILP systems (ILP:80\% and ILP:$\infty$) obtained similar results to BM13 for the French corpus.
The POS-LM post-processing further improved the ROUGE scores for Portuguese and Spanish, but unfortunately not for the French corpus.

\begin{table*}[ht]
\begin{center}
\caption{\label{rouge} ROUGE F-measure results on the French, Portuguese and Spanish corpora. The best ROUGE results are in bold.
}
\begin{tabular}{lcc:cc:cc}
\toprule
\multirow{2}{*}{\centering \bf Metrics} & \multicolumn{6}{c}{ \bf Methods} \\
 & F10 & BM13 & ILP:80\% & ILP:80\%+LM & ILP:$\infty$ & ILP:$\infty$+LM \\
 \cmidrule{1-7}
 \textbf{French} & & & & & & \\
 ROUGE-1 	& 0.6384 & 0.6674 & 0.6630 & 0.6418 & \textbf{0.6730} & 0.6460 \\
 ROUGE-2 	& 0.4423 & \textbf{0.4672} & 0.4487 & 0.4187 & 0.4567 & 0.4179 \\
 ROUGE-SU4  & 0.4297 & \textbf{0.4602} & 0.4410 & 0.4152 & 0.4511 & 0.4136 \\
 \cmidrule{1-7}
 \textbf{Portuguese} & & & & & & \\
 ROUGE-1 	& 0.5388 & 0.5532 & 0.5668 & 0.5763 & 0.5700 & \textbf{0.5811} \\
 ROUGE-2 	& 0.2971 & 0.3029 & 0.3105 & 0.3112 & 0.3132 & \textbf{0.3249} \\
 ROUGE-SU4  & 0.2938 & 0.2868 & 0.3060 & 0.3149 & 0.3057 & \textbf{0.3210} \\
 \cmidrule{1-7}
 \textbf{Spanish} & & & & & & \\
 ROUGE-1 	& 0.5004 & 0.5140 & 0.5422 & \textbf{0.5500} & 0.5425 & 0.5442 \\
 ROUGE-2 	& 0.2983 & 0.2960 & 0.3128 & \textbf{0.3195} & 0.3109 & 0.3194 \\
 ROUGE-SU4  & 0.2847 & 0.2801 & 0.2973 & \textbf{0.3052} & 0.2963 & 0.3047
 \\
 \bottomrule
\end{tabular}
\end{center}
\end{table*}

Table~\ref{size_keywords} displays the average length, the compression ratio and the average number of keywords that are kept in the final compression. 
F10 generated the shortest compressions for all corpora, our approach producing outputs of an intermediate length with respect to BM13, except for the French corpus for which ILP:$\infty$ generated slightly longer compressions.
The keyword bonus and the POS-LM score act differently on the selection of words. On the one hand, the keyword bonus promotes the integration of keywords from difference sentences. On the other hand, the POS-LM favors grammaticality and longer subsequences of the original sentences, which reduces the mix of sentences and, consequently, the number of keywords in the compressions.

\begin{table*}[ht]
\begin{center}
\caption{\label{size_keywords} Compression length (\#words), standard deviation and number of used keywords computed on the French, Portuguese and Spanish corpora.}
\begin{tabular}{lcc:cc:cc}
\toprule
\multirow{2}{*}{\centering \bf Metrics} & \multicolumn{6}{c}{ \bf Methods} \\
 & F10 & BM13 & ILP:80\% & ILP:80\%+LM & ILP:$\infty$ & ILP:$\infty$+LM \\
 \cmidrule{1-7}
 \textbf{French} & & & & & & \\
 Avg. Length     & 16.9 $\pm$ 5.1 & 19.7 $\pm$ 6.9 & 19.8 $\pm$ 4.8 & 19.5 $\pm$ 4.9 & 20.6 $\pm$ 5.5 & 20.8 $\pm$ 5.8 \\
 Comp. Ratio. (\%)   & 51.3 & 59.8 & 59.9 & 59.2 & 62.6 & 63.1 \\
 Keywords 		 & 6.8 & 7.7 & 8.3 & 7.9 & 8.5 & 8.1 \\
 \cmidrule{1-7}
 \textbf{Portuguese} & & & & & & \\
 Avg. Length 	 & 17.3 $\pm$ 5.3 & 22.9 $\pm$ 6.3 & 19.5 $\pm$ 4.0 & 19.4 $\pm$ 4.4 & 21.8 $\pm$ 5.5 & 20.5 $\pm$ 5.0 \\
 Comp. Ratio. (\%)   & 52.2 & 69.1 & 59.0 & 58.7 & 65.9 & 62.2 \\
 Keywords 		 & 7.0 & 8.5 & 8.2 & 8.0 & 8.9 & 8.3 \\
 \cmidrule{1-7}
 \textbf{Spanish} & & & & & & \\
 Avg. Length 	 & 16.5 $\pm$ 6.4 & 23.4 $\pm$ 8.4 & 20.3 $\pm$ 5.9 & 20.9 $\pm$ 5.2 & 21.1 $\pm$ 7.0 & 23.4 $\pm$ 7.3 \\
 Comp. Ratio. (\%)   & 43.2 & 61.2 & 53.2 & 54.7 & 55.4 & 61.2 \\
 Keywords 		 & 5.8 & 6.9 & 7.7 & 7.6 & 7.9 & 7.9 \\
 \bottomrule
\end{tabular}
\end{center}
\end{table*}

We also led a manual evaluation to study the informativeness and grammaticality of compressions.
We measured the inter-rater agreement on the judgments we collected, obtaining values of Fleiss' kappa of 0.423, 0.289 and 0.344 for French, Portuguese and Spanish respectively.
These results show that human evaluation is rather subjective. Questioning evaluators on how they proceed to rate sentences reveals that they often made their choice by comparing outputs for a given cluster.

Table \ref{manual_eval} shows the manual analysis that ratifies the good results of our system. Informativeness scores are consistently improved by the ILP method, whereas grammaticality results measured on the three systems are similar. Besides, statistical tests show that this enhancement regarding informativeness and grammaticality is significant for Spanish corpus.
For the Portuguese and Spanish corpora, our method obtained the best results for informativeness and grammaticality with shorter compressions.
For the French corpus, F10 obtained the highest value for grammatical quality, while BM13 generated more informative compressions.
Finally, the reranking method proposed by BM13 based on the analysis of \textit{key phrases} of candidate compression improves informativeness, but not to the same degree as our ILP model.
This more moderate enhancement can be related to the limitation of this reranking method to candidate sentences generated by F10. 

\begin{table*}[h]
\begin{center}
\caption{\label{manual_eval} Manual evaluation of compression (ratings are expressed on a scale of 0 to 2). 
The best results are in bold ($^\star$ and $^{\star\star}$ indicate significance at the 0.01 and the 0.001 level using ANOVA's test related to F10, respectively; $^\dagger$ and $^{\dagger\dagger}$ indicate significance at the 0.01 and the 0.001 level using ANOVA’s test related to BM13, respectively).}
\scalebox{0.77}{
\begin{tabular}{lcc:cc:cc}
\toprule
\multirow{2}{*}{\centering \bf Metrics} & \multicolumn{6}{c}{ \bf Methods} \\
 & F10 & BM13 & ILP:80\% & ILP:80\%+LM & ILP:$\infty$ & ILP:$\infty$+LM \\
 \cmidrule{1-7}
 \textbf{French} & & & & & & \\
 \multicolumn{2}{l}{Informativeness}  & & & & & \\
 Score 0  & 20\% & 10\% & 14\% & 16\% & 14\% & 14\% \\
 Score 1  & 36\% & 31\% & 32\% & 35\% & 27\% & 34\% \\
 Score 2  & 44\% & 59\% & 54\% & 49\% & 59\% & 52\% \\
 Avg.     &  1.25 $\pm$ 0.8  &  \textbf{1.48 $\pm$ 0.7}  &  1.40 $\pm$ 0.7  &  1.33 $\pm$ 0.7  &  1.45 $\pm$ 0.7  &  1.39 $\pm$ 0.7  \\
 \multicolumn{2}{l}{Grammaticality}  & & & & & \\
 Score 0  & 6\%  & 7\%  & 12\% & 8\%  & 10\% & 10\% \\
 Score 1  & 23\% & 29\% & 36\% & 29\% & 35\% & 36\% \\
 Score 2  & 71\% & 64\% & 52\% & 63\% & 55\% & 54\% \\
 Avg.     &  \textbf{1.65 $\pm$ 0.6}  &  1.56 $\pm$ 0.6  &  1.44 $\pm$ 0.7  &  1.55 $\pm$ 0.6  &  1.45 $\pm$ 0.7  &  1.44 $\pm$ 0.7  \\
\cmidrule{1-7}
 \textbf{Portuguese} & & & & & & \\
 \multicolumn{2}{l}{Informativeness}  & & & & & \\
 Score 0  & 9\%  & 7\%  & 8\%  & 5\%  & 7\%  & 8\% \\
 Score 1  & 30\% & 16\% & 18\% & 22\% & 12\% & 13\% \\
 Score 2  & 61\% & 77\% & 74\% & 73\% & 81\% & 79\% \\
 Avg.     &  1.51 $\pm$ 0.7 &  1.70 $\pm$ 0.6 &  1.66 $\pm$ 0.6  &  1.68 $\pm$ 0.6  &  \textbf{1.74 $\pm$ 0.6}  &  1.71 $\pm$ 0.6  \\
 \multicolumn{2}{l}{Grammaticality}  & & & & & \\
 Score 0  & 9\%  & 8\%  & 6\%  & 5\%  & 4\%  & 7\% \\
 Score 1  & 21\% & 18\% & 18\% & 21\% & 15\% & 17\% \\
 Score 2  & 70\% & 74\% & 76\% & 74\% & 81\% & 76\% \\
 Avg.     &  1.61 $\pm$ 0.6  &  1.66 $\pm$ 0.6  &  1.71 $\pm$ 0.6  &  1.69 $\pm$ 0.6  &  \textbf{1.76 $\pm$ 0.5}  &  1.68 $\pm$ 0.6  \\
 \cmidrule{1-7}
 \textbf{Spanish} & & & & & & \\
 \multicolumn{2}{l}{Informativeness}  & & & & & \\
 Score 0  & 24\% & 26\% & 12\% & 11\% & 10\% & 10\% \\
 Score 1  & 49\% & 31\% & 39\% & 36\% & 39\% & 29\% \\
 Score 2  & 27\% & 43\% & 49\% & 53\% & 51\% & 61\% \\
 Avg.     &  1.02 $\pm$ 0.7  &  1.16 $\pm$ 0.8  &  1.36 $\pm$ 0.7 $^{\star\star}$ &  1.41 $\pm$ 0.7 $^{\star\star}$ &  1.40 $\pm$ 0.7 $^{\star\star}$ &  \textbf{1.50 $\pm$ 0.7} $^{\star\star\dagger\dagger}$ \\
 \multicolumn{2}{l}{Grammaticality}  & & & & & \\
 Score 0  & 11\% & 18\% & 12\% & 8\%  & 10\% & 6\% \\
 Score 1  & 26\% & 33\% & 35\% & 36\% & 35\% & 29\% \\
 Score 2  & 63\% & 49\% & 53\% & 56\% & 55\% & 65\% \\
 Avg.     &  1.51 $\pm$ 0.7  &  1.30 $\pm$ 0.8  &  1.40 $\pm$ 0.7  &  1.48 $\pm$ 0.6  &  1.45 $\pm$ 0.7  &  \textbf{1.59 $\pm$ 0.6} $^{\dagger}$ \\
 \bottomrule
\end{tabular}
}
\end{center}
\end{table*}

\subsection{Discussion}

Short compressed sentences are appropriate to summarize documents; however, they may remove key information and prejudice the informativeness of the compression.
For instance, for the sentences that would be associated with a higher relevant score by the \ac{ATS} system, producing longer sentences would be more appropriate. 
Generating longer sentences makes easier to keep informativeness but often increases difficulties to have a good grammatical quality while combining different parts of sentences.
Depending on the kind of cluster short compressions can be generated or not with good informativeness scores. In that respect, the system has to adapt its analysis to generate long or short sentences.

F10 produced the shortest compressions for all corpora but its outputs have the worst informativeness score.
BM13 improved these results; however, their compressions are longer than F10 (for all corpora) and our system (for the Portuguese and the Spanish corpora).
For Spanish, the informativeness scores of all versions of our method are statistically better than F10, and the version ILP:$\infty$+LM is statistically better than both baselines for this corpus.
Given the small difference of informativeness between BM13 and our ILP approach for the Portuguese and the French corpora, we analyzed the relation between informativeness and \ac{CR} to define which method obtains the best results.
For Portuguese, BM13 and all versions of our system achieved similar informativeness scores, whereas our method generated significantly shorter compressions with an absolute decrease in the range 3.0--10.1 points.
For the French corpus, it is complicated to define the best system because the second baseline, ILP:80\% and ILP:$\infty$ have similar informativeness scores for similar \ac{CR}s.
An inspection of the compressions generated by all systems highlighted that the low performance of our approach for the French dataset is partly related to the structure of negative sentences in French. 
In this language, these sentences must usually be composed of the tokens ``ne'' and ``pas'' to be correct, like in the following example: ``La France n'a pas remporté le championnat du monde de volley-ball'' (France did not win the world volleyball championship). In the studied dataset, the French corpus contains 27 negative source sentences divided into 13 clusters.
Our approach often missed one of these tokens in its output compressions with the negative structure, which reduced the scores for informativeness and grammaticality.
A post-processing of compressions could check if these two tokens are presented in the compression and correct this error.

Tables~\ref{size_keywords} and \ref{manual_eval} show that the informativeness scores and keywords are related, i.e., the higher the number of keywords the higher the informativeness score.
According to its type (with respect to the size and the amount of information), a cluster can have a different number of real keywords (more or less than 10 keywords).
The number of keywords and informativeness scores are related, except for BM13 on the French corpus that used fewer keywords than our method and still generated more informative compressions.

The POS-LM post-processing does not improve significantly the compression quality of our method.
This post-processing maintain or enhance grammaticality for all corpora, except for the ILP:$\infty$+LM for Portuguese corpus, and informativeness for the Portuguese and the Spanish corpora.
The biggest difference between these two versions of all methods is on the Spanish corpus (differences of 0.1 and 0.14 are observed for informativeness and grammaticality, respectively), for which the POS-LM version generated a longer version (\ac{CR} is increased by 5.8 points), which justifies the improvement of informativeness.

\subsection{Applications}

Most of previous MSC approaches have been applied on the Text Summarization problem and its variations.
Among these works, several versions of our ILP method on different types of documents and in multiple languages have been successfully tested.

In the first application, the ILP approach was applied to the problem of microblog contextualization~\cite{linhares:2018:clef,linhares:2017:clef}. Given a microblog about a festival, Linhares Pontes et al.'s ~\cite{linhares:2018:clef,linhares:2017:clef} system was able to generate a summary (maximum of 120 words) in four languages (English, French, Portuguese and Spanish) of Wikipedia’s pages describing this microblog. In order to get more information about these festivals, 
they used Wikipedia to find information about these festivals and adapt the MSC method to extract relevant information related to the festival and generate a summary. 

Linhares Pontes et al.~\cite{linhares:2018:missi} also investigated the generation of cross-lingual speech summaries of news documents. The goal was to analyze an audio file in French and generate a text summary in English. Contrary to the text document, the transcription of audio files must use Automatic Speech Recognition (ASR), which complicates and reduces the quality of the summary generation. 
They adapted the MSC method to analyze sentences, both in their original and translated forms, and generate informative compressions in English using the relevance of French and English sentences. 
Their MSC method also analyzed 3 grams to add grammatically correct sequences of words into the compressions. This feature allowed 
their method to generate compressions with a good grammaticality, even when there are erroneous transcribed sentences.

Finally, 
Linhares Pontes et al.~\cite{linhares:micai:2018,elvys:2018:nldb, LINHARESPONTES2020101763} also dealt with the issue of Cross-Language Text Summarization to generate English and French summaries from clusters of news documents in French, Portuguese and Spanish languages. 
Their MSC approach was applied on similar sentences among the documents to summarize. Despite the variety of these sentences (short, long, verbal and non-verbal sentences) and the introduction of errors by the used machine translation engine, experiments showed that the system usually generated correct compressions that are shorter and more informative than their source sentences.
\section{Conclusion}
\label{sc:conc}

Multi-Sentence Compression aims to generate a short informative text summary from several sentences with related and redundant information. Previous works built word graphs weighted by cohesion scores from the input sentences, then selected the best path to select words of the output sentence. We introduced in this study a model for \ac{MSC} with two novel features. Firstly, we extended the work done by Boudin and Morin~\cite{boudin:2013} that introduced keywords to post-process lists of N-best compressions. We proposed to represent keywords as labels directly on the vertices of word graphs to ensure the use of different keywords in the selected paths. Secondly, we devised an ILP modeling to take into account these new features with the cohesion scores, while selecting the best sentence. The compression ratio can be modulated with this modeling, by selecting for example a higher number of keywords for the sentences considered essential for a summary.

Our methodology was evaluated on three corpora built from Google news: a first one in French which had been built and used in \cite{boudin:2013}, a second and a third one in Portuguese and in Spanish~\cite{elvys:lrec}.
Automatic measures with the ROUGE package were supplemented with a manual evaluation carried out by human judges in terms of informativeness and grammaticality. We showed that keywords are important features to produce valuable compressed sentences.
The paths selected with theses features generate results consistently improved in terms of informativeness while keeping up their grammaticality.

There are several potential avenues of work. We can use other kinds of language models based on Neural Networks~\cite{Niu:2017} as an additional score to the optimization criterion to improve grammaticality. Another objective can be to manage polysemy through the use of the same label for the synonyms of each keyword inside the word graph.
Finally, \ac{MSC} can be jointly employed with the classical methods of Automatic Text Summarization by extraction in order to generate better summaries.

\section{Appendix}

Two examples in Spanish and Portuguese are provided in this section to illustrate the differences observed between the tested methods. 

\subsection{Spanish}
The Spanish cluster (Table \ref{tb:example_spanish}) is composed of 20 similar sentences. The vocabulary of this cluster is composed of 880 tokens and this cluster has a TTR of 33.3\%.
F10 generated the shortest compression; however, the sentence has missing information. The second baseline system and our method without post-processing generated incorrect compressions. Our method without post-processing generated a sentence with relevant keywords but it is not correct. The post-processing selected a more grammatical compression without reducing informativeness.
The top 10 keywords selected by LDA were : \textit{vuelo}, \textit{cuba}, \textit{fort}, \textit{lauderdale}, \textit{unidos}, \textit{primer}, \textit{jetblue}, \textit{comercial}, \textit{clara} and \textit{florida}.

\begin{table*}[h!]
\begin{center}
\caption{\label{tb:example_spanish} Example in Spanish showing the first 3 sentences among 20 source sentences and 1 of 3 available references.}
\begin{tabular}{ll}
\toprule
\multicolumn{2}{c}{\textbf{Source document}}\\
\multicolumn{2}{p{15cm}}{\small El vuelo 387 de la aerolínea estadounidense JetBlue inauguró una nueva era en el transporte entre ambos países, al partir desde Fort Lauderdale (Florida, sureste) cerca de las 10:00 locales (14H00 GMT), y llegar a Santa Clara, 280 Km al este de La Habana, a las 10:57. (\textit{Flight 387 of the US airline JetBlue inaugurated a new era in transport between the two countries, departing from Fort Lauderdale (Florida, southeast) at around 10:00 local time (14H00 GMT), and arriving in Santa Clara, 280 km east of Havana, at 10:57.})\newline
Un avión de pasajeros de la línea aérea JetBlue despegó este miércoles a Cuba desde el aeropuerto Internacional de Fort Lauderdale en lo que viene a ser el primer vuelo regular entre Estados Unidos y la isla caribeña desde 1961, en un nuevo hito en la nueva fase de relaciones entre Washington y La Habana. (\textit{A JetBlue airliner took off for Cuba on Wednesday from Fort Lauderdale International Airport, thus becoming the first regular flight between the United States and the Caribbean island since 1961, as a new milestone in the new phase of relations between Washington and Havana.})\newline
La aerolínea JetBlue inaugurará los vuelos directos comerciales el 31 de agosto con un viaje entre Fort Lauderdale, Florida, hasta el  aeropuerto de Santa Clara, a unos 270 kilómetros al este de La Habana, reportó la compañía estadunidense. (\textit{JetBlue will inaugurate direct commercial flights on Aug. 31 with a trip from Fort Lauderdale, Florida, to Santa Clara airport, some 270 kilometers east of Havana, the U.S. company reported.})
}\\
\multicolumn{2}{c}{\textbf{Reference}}\\
\multicolumn{2}{p{15cm}}{\small La aerolínea JetBlue Airways Corp inauguró el 31 de agosto los vuelos directos entre Estados Unidos y Cuba tras 50 años de suspensión . 
\textit{(The airline JetBlue Airways Corp opened on August 31 direct flights between the United States and Cuba after 50 years of suspension .)}
}
\\
\cmidrule{1-2}
\multicolumn{2}{c}{\textbf{Compressions}}\\
F10:           		 & \multicolumn{1}{p{12cm}}{\small la aerolínea \underline{jetblue} inauguró este miércoles a \underline{cuba} el \underline{primer} \underline{vuelo} inaugural . \textit{(the airline \underline{jetblue} opened the inaugural \underline{first} \underline{flight} to \underline{cuba} this wednesday .)}}\\
BM13: 			     & \multicolumn{1}{p{12cm}}{\small el aeropuerto de \underline{fort} \underline{lauderdale} , \underline{florida} , sureste de estados \underline{unidos} y \underline{cuba} desde 1961 partió este miércoles el \underline{primer} \underline{vuelo} inaugural . \textit{(the airport of \underline{fort} \underline{lauderdale} , \underline{florida} , southeastern  \underline{united} states and \underline{cuba} since 1961 departed this Wednesday on the inaugural \underline{first} \underline{flight} .)}}\\
ILP:80\%             & \multicolumn{1}{p{12cm}}{\small el aeropuerto de \underline{fort} \underline{lauderdale} , \underline{florida} , sureste de estados \underline{unidos} y \underline{cuba} desde 1961 partió este miércoles el \underline{primer} \underline{vuelo} inaugural . \textit{(the airport of \underline{fort} \underline{lauderdale} , \underline{florida} , southeastern \underline{united} states  and \underline{cuba} since 1961 departed this Wednesday on the inaugural \underline{first} \underline{flight} .)}}\\
ILP:80\%+LM          & \multicolumn{1}{p{12cm}}{\small la aerolínea \underline{jetblue} inauguró este miércoles el \underline{primer} \underline{vuelo} desde \underline{fort} \underline{lauderdale} , \underline{florida} , sureste de estados \underline{unidos} a \underline{cuba} desde 1961 . \textit{(the airline \underline{jetblue} opened Wednesday the \underline{first} \underline{flight} from \underline{fort} {lauderdale} , \underline{florida} , southeastern \underline{united} states to \underline{cuba} since 1961.)}  }\\
ILP:$\infty$         & \multicolumn{1}{p{12cm}}{\small el aeropuerto de \underline{fort} \underline{lauderdale} , \underline{florida} , sureste de estados \underline{unidos} y \underline{cuba} desde 1961 partió este miércoles el \underline{primer} \underline{vuelo} inaugural . \textit{(the airport of \underline{fort} \underline{lauderdale} , \underline{florida} , southeastern \underline{united} states and \underline{cuba} since 1961 departed this Wednesday on the inaugural \underline{first} \underline{flight} .)}}\\
ILP:$\infty$+LM      & \multicolumn{1}{p{12cm}}{\small la aerolínea \underline{jetblue} inauguró este miércoles el \underline{primer} \underline{vuelo} desde \underline{fort} \underline{lauderdale} , \underline{florida} , sureste de estados \underline{unidos} a \underline{cuba} desde 1961 . \textit{(\underline{jetblue} 
\underline{airlines} inaugurated this wednesday the \underline{first} \underline{flight} from \underline{fort} \underline{lauderdale}, \underline{florida} , southeastern \underline{united} states to \underline{cuba} since 1961 .)}}
\\\bottomrule
\end{tabular}
\end{center}
\end{table*}

\subsection{Portuguese}
Table~\ref{tb:example_portuguese} displays
a cluster composed of 11 Portuguese sentences with a TTR of 37\% and a vocabulary of 351 tokens.
In this case, F10 did not generate the shortest compression and has incorrect information. The second baseline, which post-processes the outputs of the first one, was not able to correct the errors. Almost all versions of our method generated the shortest and the most informative compressions related to the text. Our method without post-processing generated the best compression. The post-processing selected a more grammatically correct sentence, while its information is incorrect.
The top 10 keywords selected by LDA were : \textit{tesla}, \textit{solarcity}, \textit{milhões}, \textit{2,6}, \textit{solar}, \textit{empresa}, \textit{carros}, \textit{fabricante}, \textit{dólares} and \textit{motors}.

\begin{table*}[h]
\begin{center}
\caption{\label{tb:example_portuguese} Example in Portuguese showing the first 3 sentences among 11 source sentences and 1 of 2 available references.}
\begin{tabular}{ll}
\toprule
\multicolumn{2}{c}{\textbf{Source document}}\\
\multicolumn{2}{p{15cm}}{\small A Tesla fez uma oferta de compra à empresa de serviços de energia solar SolarCity por mais de 2300 milhões de dólares \textit{(Tesla made an offer to purchase the SolarCity solar energy services company for over 2,300 million dollars.)}. \newline
A Tesla Motors, fabricante de carros elétricos, anunciou aquisição da SolarCity por US\$ 2,6 bilhões \textit{(Tesla Motors, a manufacturer of electric cars, announced the purchase of SolarCity for \$2.6 billion.)}. \newline
A fabricante de carros elétricos e baterias Tesla Motors disse nesta segunda-feira (1) que chegou a um acordo com a SolarCity para comprar a instaladora de painéis solares por US\$ 2,6 bilhões, em um grande passo do bilionário Elon Musk para oferecer aos consumidores um negócio totalmente especializado em energia limpa, informou a Reuters (\textit{Electric car and battery manufacturer Tesla Motors said on Monday (1) that it reached an agreement with SolarCity to buy the solar panel installer for \$2.6 billion, in a big step took by billionaire Elon Musk to offer consumers a fully specialized clean energy business, Reuters reported.}). 
}\\
\multicolumn{2}{c}{\textbf{Reference}}\\
\multicolumn{2}{p{15cm}}{\small A Tesla Motors anunciou acordo para comprar a SolarCity por US\$ 2,6 bilhões.
(\textit{Tesla Motors has announced an agreement to buy SolarCity for US\$ 2.6 billion.})}
\\
\cmidrule{1-2}
\multicolumn{2}{c}{\textbf{Compressions}}\\
\multirow{ 2}{*}{F10}					 & \multicolumn{1}{p{12cm}}{\small a \underline{solarcity} para comprar a instaladora de painéis \underline{solares} por us\$ \underline{2,6} bilhões\textit{ ( \underline{solarcity} to buy the \underline{solar} panel installer for us\$ \underline{2.6} \underline{billions} .)}} \\
\multirow{ 2}{*}{BM13}			     & \multicolumn{1}{p{12cm}}{\small a \underline{solarcity} para comprar a instaladora de painéis \underline{solares} por us\$ \underline{2,6} mil \underline{milhões} de dólares \textit{(\underline{solarcity} to buy the \underline{solar} panel installer for us\$ \underline{2.6} \underline{billion} dollars).}} \\
\multirow{ 2}{*}{ILP:80\%}             & \multicolumn{1}{p{12cm}}{\small a \underline{tesla} vai comprar a \underline{solar} \underline{solarcity} por \underline{2,6} mil \underline{milhões} de dólares \textit{(\underline{tesla} will buy the \underline{solar} \underline{solarcity} for \underline{2.6} \underline{billion} dollars.)}} \\
\multirow{ 2}{*}{ILP:80\%+LM}          & \multicolumn{1}{p{12.1cm}}{\small a \underline{solarcity} para comprar a instaladora de painéis \underline{solares} por \underline{2,6} mil \underline{milhões} de dólares \textit{(\underline{solarcity} to buy the \underline{solar} panel installer for \underline{2.6} \underline{billion} dollars.)}} \\
\multirow{ 2}{*}{ILP:$\infty$}         & \multicolumn{1}{p{12cm}}{\small a \underline{tesla} vai comprar a \underline{solar} \underline{solarcity} por \underline{2,6} mil \underline{milhões} de dólares \textit{(\underline{tesla} will buy the \underline{solar} \underline{solarcity} for \underline{2.6} \underline{billion} dollars.)}} \\
\multirow{ 2}{*}{ILP:$\infty$+LM}      & \multicolumn{1}{p{12.1cm}}{\small a \underline{solarcity} para comprar a instaladora de painéis \underline{solares} por \underline{2,6} mil \underline{milhões} de dólares \textit{(\underline{solarcity} to buy the \underline{solar} panel installer for \underline{2.6} \underline{billion} dollars.)}} \\
\bottomrule
\end{tabular}
\end{center}
\end{table*}

\section*{Acknowledgments}
This work was partially financed by the European Project CHISTERA-AMIS ANR-15-CHR2-0001 and the European Union's Horizon 2020 research and innovation program under grants 770299 (NewsEye) and 825153 (EMBEDDIA).

\small{
\bibliographystyle{cys}
\bibliography{references}
}


\end{document}